\documentclass[runningheads]{llncs}

\usepackage[T1]{fontenc}
\usepackage{graphicx}
\usepackage{color}
\usepackage[colorlinks=true, urlcolor=blue]{hyperref}

\usepackage{booktabs}
\usepackage{subfigure}
\usepackage{amsmath}
\usepackage{amssymb}
\usepackage{booktabs}
\usepackage{multirow}

\usepackage{float}

\begin{document}

\title{Planner-Conditioned Diffusion for Coordinated Multi-Agent Exploration}

\titlerunning{Planner-Conditioned Diffusion for Multi-Agent Exploration}

\author{
Marcus Yu Siong Teo\thanks{The first two authors contributed equally to this work.}
\and
Jeric Lew$^{\star}$
\and
Tanishq Duhan
\and
Guillaume Sartoretti
}

\authorrunning{M. Y. S. Teo et al.}

\institute{
Department of Mechanical Engineering, National University of Singapore, Singapore
\email{\{e0958027,jericlew,e1280621\}@u.nus.edu,
mpegas@nus.edu.sg}
\url{https://marmotlab.org/}
}

\maketitle

\begin{abstract}
Coordinated multi-agent exploration requires not only efficient individual coverage but also non-redundant coverage across agents over extended planning horizons. Conventional approaches rely on hand-crafted coordination rules, while end-to-end multi-agent learning methods are difficult to scale and train. Diffusion-based planners such as DARE offer a promising alternative by generating long-horizon trajectories instead of single-step actions, but existing methods are trained on a narrow planner distribution, limiting behavioral diversity and inference-time controllability.
We propose a Planner-Conditioned Diffusion Policy (PCDP) for graph-based multi-agent exploration. PCDP is trained on demonstrations from multiple planner styles with planner identity as an explicit conditioning input, enabling a single shared model to learn a multimodal trajectory distribution and generate diverse, controllable trajectory candidates from the same observation. Rather than learning coordination end-to-end, we reuse this multimodal single-agent policy across all agents and introduce coordination through local reranking, in which nearby agents jointly select the trajectory combination with minimal predicted overlap.
We evaluate PCDP against classical and diffusion-based baselines on 100 held-out maps in a four-agent simulation setting. PCDP matches the perfect success rate of the diffusion-based baselines while improving mean max-agent travel by 1.94\%, mean total team travel by 1.58\%, and mean agent imbalance by 5.66\%. Crucially, reranking alone over a single-planner baseline yields only marginal gains, indicating that planner-conditioned multimodality is the main contributor to improved coordination. Qualitative simulation results and real-robot experiments with two agents further validate that diverse long-horizon trajectory generation produces emergent spatial separation between agents without any explicit repulsion mechanism. Code available at \href{https://github.com/marmotlab/PCDP}{github.com/marmotlab/PCDP}.

    \keywords{Multi-agent systems \and Robot exploration \and Diffusion Policy}
\end{abstract}

\section{Introduction}

\begin{figure}[t]
    \centering
    \includegraphics[width=\textwidth]{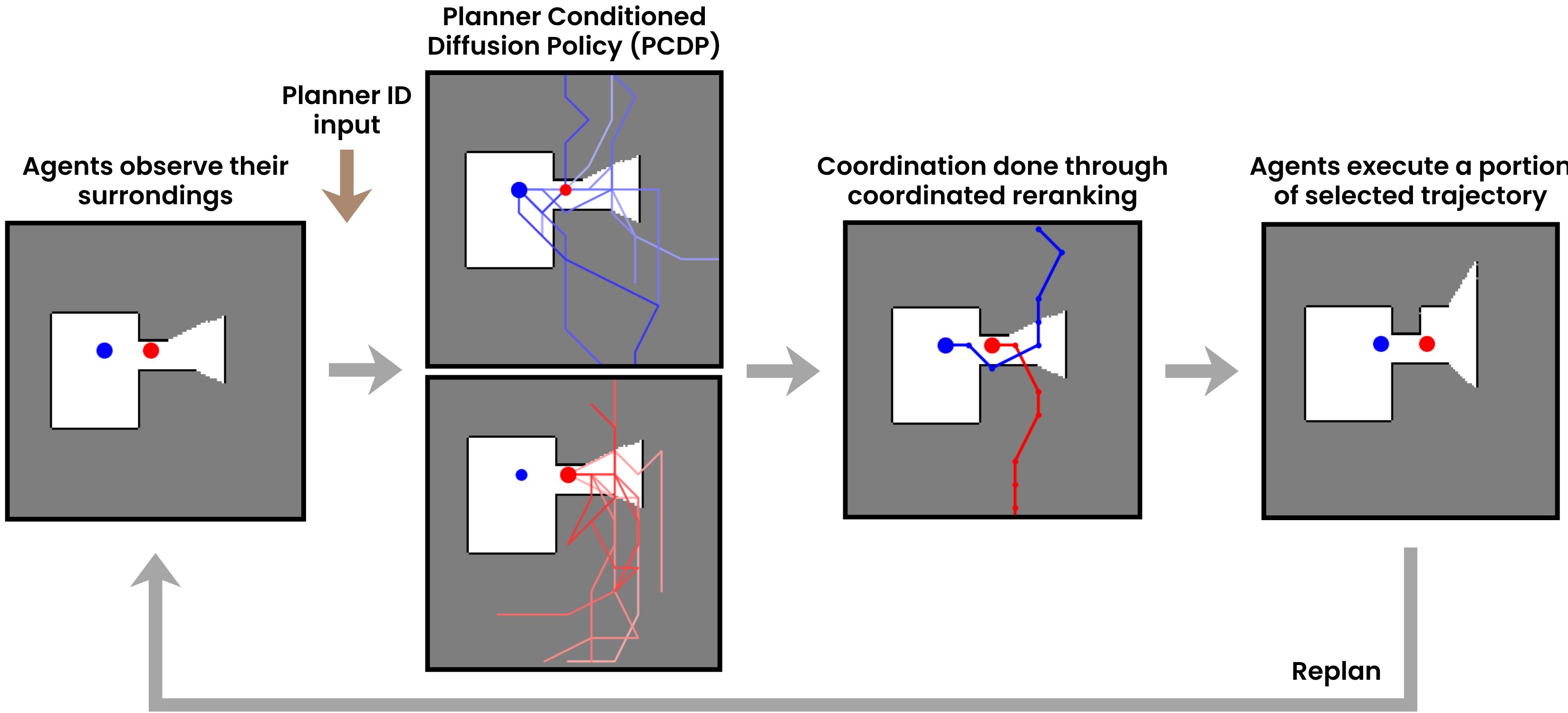}
    \caption{PCDP execution with two agents. Each agent independently samples a diverse set of candidate trajectories from the diffusion policy, conditioned on its own graph observation and multiple planner identities to maximize behavioral diversity. Agents within coordination range perform local reranking over their joint candidate sets, selecting the combination that minimizes predicted overlap and redundant coverage. The selected trajectory is then executed in a receding-horizon manner before replanning.}
    \label{fig:intro_example}
    \vspace{-1.5em}
\end{figure}

Autonomous exploration is a well-researched problem in robotics, with applications in search and rescue, infrastructure inspection, and mapping of unknown environments. The objective is to uncover unseen areas while maximizing efficiency in terms of distance traveled or time taken. This requires reasoning over partial observations and making sequential decisions under uncertainty, which makes exploration inherently challenging even in the single-agent setting~\cite{cao2023ariadne,yamauchi1997frontier}.

A natural extension is to deploy multiple agents to accelerate coverage in large or time-critical environments. However, multi-agent exploration introduces additional complexity: agents must not only explore efficiently individually, but also coordinate to avoid redundant coverage and ensure effective spatial distribution~\cite{burgard2005coordinated,yamauchi1998frontier}. The quality of the team's performance depends on how well agents' behaviors complement each other over time.

Classical frontier-based, information-theoretic, next-best-view, and hierarchical methods have achieved strong practical results through interpretable geometric objectives~\cite{bircher2016receding,cao2021tare,gao2018improved,yamauchi1997frontier}. Their behavior, however, is largely determined by the objective or heuristic specified by the designer, motivating complementary learned models that can represent long-horizon exploration behaviors.

Learning-based methods offer a data-driven alternative, with learned policies selecting next actions or local goals~\cite{cao2023ariadne,cao2024deep}. End-to-end multi-agent learning approaches attempt to address coordination directly~\cite{yang2023active,zhu2024maexp}, but such training is often computationally expensive, difficult to scale, and sensitive to training instability. Diffusion-based planners for exploration~\cite{chi2024diffusionpolicy,ho2020denoisingdiffusionprobabilisticmodels,janner2022diffuser}, such as DARE~\cite{cao2025dare}, provide a promising direction by representing distributions over complete trajectories and supporting multimodal long-horizon generation, enabling richer reasoning over future motion. However, existing diffusion-based exploration methods are typically trained on a single planner distribution, resulting in limited behavioral diversity and reduced controllability at inference time.

In this work, we aim to leverage long-horizon diffusion-based planning for multi-agent exploration without requiring end-to-end multi-agent training. We propose a Planner-Conditioned Diffusion Policy (PCDP), as shown in Figure~\ref{fig:intro_example}, that is trained on trajectories from multiple planner types, with planner identity provided as an explicit conditioning variable. This enables the model to learn a multimodal distribution over different exploration behaviors and to generate diverse trajectory candidates under different planner conditions. At inference time, we reuse this shared single-agent policy across all agents and introduce coordination through a lightweight, local reranking procedure that selects complementary trajectories with minimal overlap.

Our key insight is that adding structured diversity to long-horizon trajectory candidates directly improves multi-agent coordination under a fixed selection mechanism. We evaluate PCDP against classical and diffusion-based baselines on 100 held-out maps in a four-agent simulation setting. PCDP + local reranking achieves full exploration coverage and the lowest mean travel and agent imbalance among all compared methods. We further validate the approach on a real two-agent robot platform, demonstrating that our proposed coordination behavior transfers to physical deployment. The contributions of this work are: (1) a planner-conditioned diffusion policy that learns a controllable, multimodal trajectory distribution for exploration; (2) a modular multi-agent framework that achieves coordination via a shared trajectory generator reused across agents with local-range inference-time reranking; and (3) empirical evidence that enhanced trajectory diversity improves coordination efficiency without end-to-end multi-agent training, validated in both simulation and real-robot experiments.

\section{Related Work}

Classical exploration has long been studied through frontier-based and information-driven formulations. Frontier-based exploration remains a foundational approach for directing an agent toward the boundary between known and unknown space~\cite{yamauchi1997frontier}. Subsequent work has extended this line with stronger heuristics, information-gain objectives, and next-best-view style criteria~\cite{bircher2016receding,gao2018improved}, with TARE~\cite{cao2021tare} in particular representing a strong hierarchical classical planner that serves as a key demonstration source in our dataset. In multi-agent settings, these approaches are often combined with explicit coordination logic to reduce redundant exploration and improve coverage efficiency~\cite{burgard2005coordinated,yamauchi1998frontier}. Our approach is complementary to this literature: rather than introducing a new frontier utility or assignment rule, we represent multiple planner-inspired behaviors within a shared learned trajectory generator.

Learning-based exploration methods~\cite{cao2023ariadne,cao2024deep,li2019deep,xu2022explore,zhu2018deep} learn policies directly from data or interaction. End-to-end multi-agent learning approaches~\cite{chiun2025marvel,zhu2024maexp} address coordination directly by jointly training agents, but are often computationally expensive, difficult to scale, and sensitive to parameter tuning. More recent work has also explored topological and active mapping formulations for multi-agent exploration~\cite{lee2024multi,yang2023active}. However, many of these methods still operate at the level of action prediction or local decision-making, rather than directly modeling a distribution over future trajectories. In contrast, our approach shares a single-agent policy across all agents and introduces coordination entirely at inference time, avoiding joint training while placing greater demand on the quality and diversity of trajectory candidates.

Diffusion models are increasingly used for trajectory generation in robotics, as they can represent multimodal trajectory distributions and support long-horizon planning through iterative denoising~\cite{chi2024diffusionpolicy,janner2022diffuser}. DARE applies diffusion-based trajectory generation to autonomous exploration by learning long-term exploration trajectories from planner demonstrations~\cite{cao2025dare}. PCDP extends this direction by training on a planner-diverse dataset rather than a narrow planner set, and by treating planner identity as a controllable conditioning signal, allowing the same observation to generate behaviorally distinct trajectory modes. These planner-conditioned trajectories then serve directly as the candidate source for coordinated multi-agent planning through inference-time reranking.

\section{Background}

\subsection{Multi-Agent Exploration}

We consider $n$ agents exploring a bounded and initially unknown environment represented as a 2D occupancy grid map $\mathcal{M}$. The map is partitioned into free, occupied, and unknown regions $\mathcal{M}_f$, $\mathcal{M}_o$, and $\mathcal{M}_u$, respectively, with $\mathcal{M}_f \cup \mathcal{M}_o \cup \mathcal{M}_u=\mathcal{M}$. Agents operate under global communication and plan from a shared map updated with all observations. Exploration is complete when $\mathcal{M}_u=\emptyset$.
 
At each decision step $t$, each agent $i$ selects and executes an action $a_t^{(i)}$. The team objective is to minimize the maximum agent path length required to complete exploration:
\begin{equation}
    \Psi^* = \arg\min_{\Psi}\, \max_{i \in [1,n]}\, L\!\left(\psi^{(i)}\right),
    \quad \text{s.t.} \quad \mathcal{M}_u = \emptyset
\end{equation}
where $L(\psi^{(i)})$ denotes the total path length of agent $i$ and $\Psi = \{\psi^{(1)}, \dots, \psi^{(n)}\}$ is the set of agent trajectories.

\subsection{Diffusion-Based Exploration Planning}
\label{sec:DARE}

Diffusion models~\cite{ho2020denoisingdiffusionprobabilisticmodels} generate samples by reversing a forward process that progressively corrupts a clean sample $x_0$ with Gaussian noise. At inference, the learned reverse process denoises iteratively from a Gaussian prior conditioned on available context, producing diverse outputs across repeated samples.

DARE~\cite{cao2025dare} applies this framework to single-agent exploration planning, training a conditional diffusion policy $p_\theta(\tau_t \mid O_t)$ on demonstrations from a ground-truth planner with full map access. Here $\tau_t$ is the window of actions (trajectory) generated at planning step $t$ defined in Eq.~\eqref{eq:tau} below, $O_t = (o_{t-T_o+1}, \dots, o_t)$ is the observation context formed by the $T_o$ most recent graph observations, $p_\theta$ is the conditional trajectory distribution induced by the learned reverse process given $O_t$, and $\theta$ contains all learnable policy parameters. Exploration is performed on a collision-free uniform graph $\mathcal{G}_t = (\mathcal{V}_t, \mathcal{E}_t)$, where nodes $\mathcal{V}_t$ are placed at regular intervals of $d_n$ meters in the known free space and edges $\mathcal{E}_t$ connect nodes with collision-free line-of-sight paths. At each step, the agent receives a graph-based observation $o_t$ in which each node encodes its position relative to the agent $(x, y)$, a utility value counting reachable frontier cells, and a binary guidepost signal toward the nearest frontier. This representation gives the agent a structured, global view of where unexplored area remains and how to navigate toward it. The diffusion model then outputs an action sequence of future node positions with the same grid layout as the uniform graph.

The policy operates over three horizons. The observation horizon $T_o$ is the number of recent observations forming the conditioning context $O_t$, with the corresponding past actions retained in the generated window. The prediction horizon $T_p$ is the length of the window produced by a single denoising pass, so $\tau_t$ contains exactly $T_p$ elements:
\begin{equation}
    \label{eq:tau}
    \tau_t = \left(a_{t-T_o+1},\, \dots,\, a_t,\, \dots,\, a_{t-T_o+T_p}\right)
\end{equation}
of which the first $T_o-1$ are past actions, so the action taken from the current observation sits at index $T_o-1$. The action horizon $T_a$ is the number of actions executed before replanning. DARE follows a receding horizon strategy with $T_a = 1$: only $\tau_t[T_o-1] = a_t$ is executed before the shared map is updated and the policy replans, allowing for long-horizon reasoning and reactivity to new information.

Because the denoising process is stochastic, repeated sampling from the same observation produces a diverse set of trajectory candidates, each representing a plausible future path as shown in Figure~\ref{fig:intro_example}. This property makes diffusion policies naturally suited for candidate-based planning, where multiple trajectories are generated and the best is selected. However, single-planner demonstration data may concentrate these samples around a dominant behavior, limiting the diversity available to downstream trajectory selection.

\section{Method}

\begin{figure}[t]
    \centering
    \includegraphics[
        width=\textwidth,
        trim={0 2.8cm 0 4cm},
        clip,
        keepaspectratio
    ]{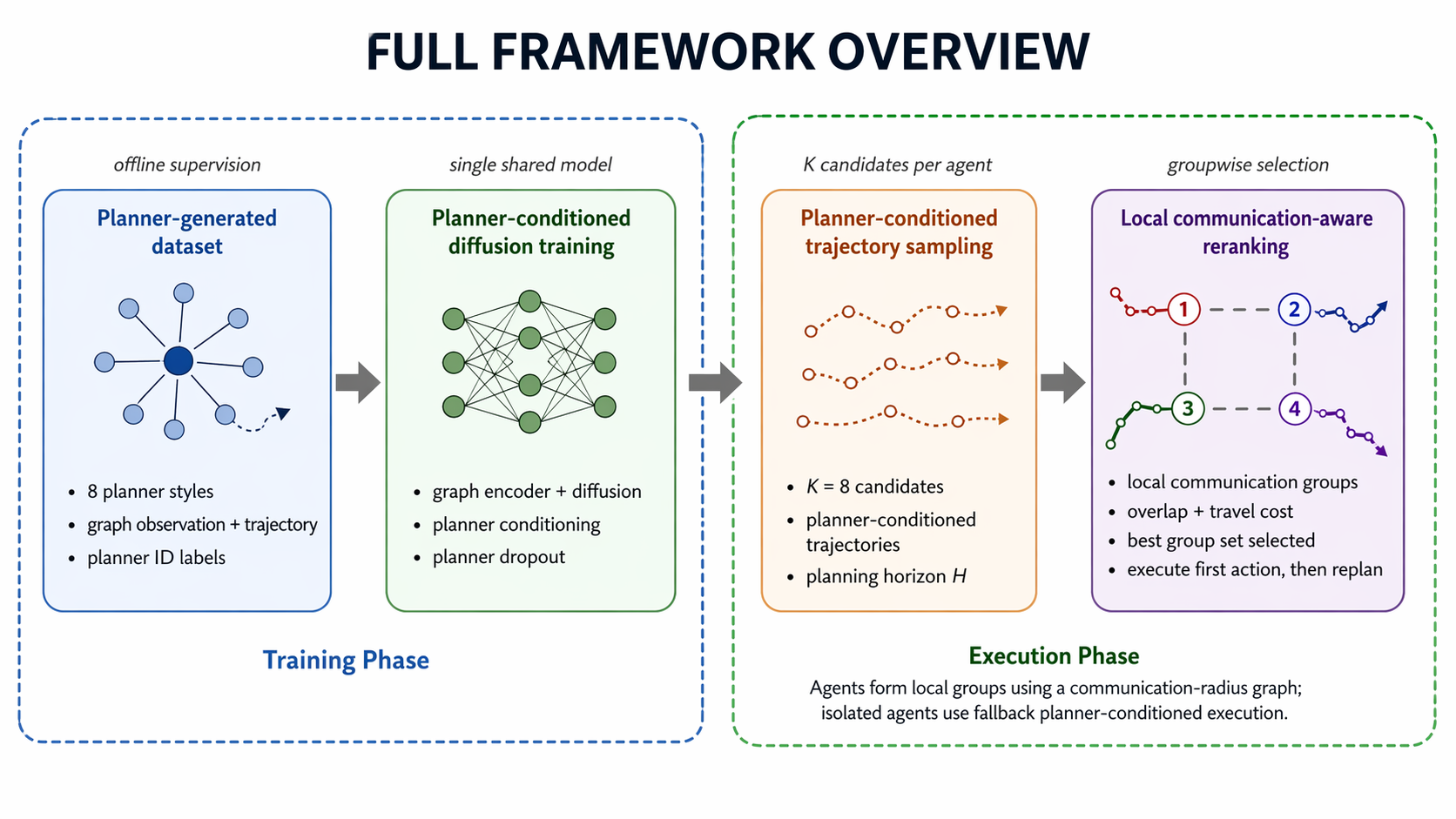}
    \caption{PCDP training and execution pipeline. Planner-diverse demonstrations train a diffusion policy conditioned on graph observation and planner identity; at inference, locally coordinating agents jointly rerank sampled trajectories.}
    \label{fig:overview}
\end{figure}

Figure~\ref{fig:overview} summarizes the three components of PCDP: planner-diverse dataset construction, planner-conditioned diffusion policy training, and inference-time local reranking for multi-agent coordination. The learning problem is kept at the single-agent level throughout; coordinated behavior emerges from trajectory selection rather than joint training.

\subsection{Planner-Diverse Demonstration Dataset}

\begin{figure}[!t]
    \centering
    \includegraphics[
        width=\textwidth,
        trim={0 1.8cm 0 1.8cm},
        clip,
        keepaspectratio
    ]{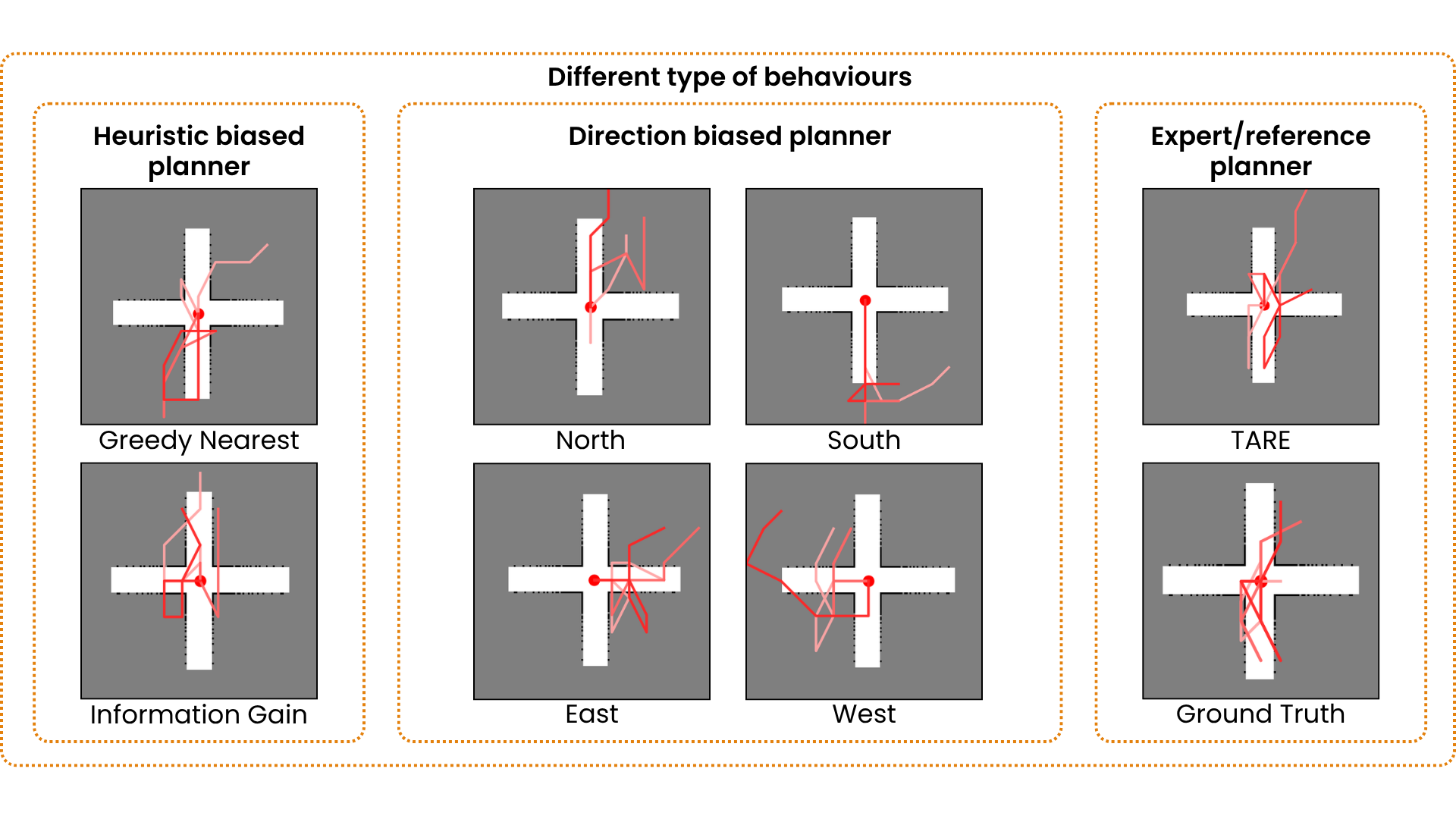}
    \caption{Trajectory diversity under planner conditioning. From similar graph observations, different planner conditions produce distinct trajectory modes, illustrating that the learned policy captures a multimodal exploration distribution.}
    \label{fig:trajectory_diversity}
\end{figure}

To broaden the behavioral diversity of the trajectory generator, we construct demonstrations from $M = 8$ planner classes selected to provide complementary exploration preferences: Greedy Nearest favors short-range frontier travel, Information Gain prioritizes high-utility regions, four directional-bias planners induce spatially distinct motion, TARE~\cite{cao2021tare} contributes hierarchical long-range behavior, and the Ground-Truth planner provides privileged expert-style coverage, as shown in Figure~\ref{fig:trajectory_diversity}. Together these span distance-, information-, direction-, and expert-driven behaviors.

Each training sample takes the form $(O_t,\, \tau_t,\, c_t)$, where $O_t$ is the observation context at decision time $t$, $\tau_t$ is the trajectory produced by the selected planner, and $c_t$ is the associated planner label. The purpose of planner diversity is not merely to increase sample count, but to expose the model to structurally distinct exploration behaviors. Planner identity therefore acts as a structured source of behavioral variation rather than generic data augmentation. The baseline diffusion model follows the narrower DARE-style training setup~\cite{cao2025dare}, using demonstrations from a single ground-truth planner.

\subsection{Planner-Conditioned Diffusion Policy}

We extend the DARE baseline by introducing planner identity as an explicit conditioning variable.
Rather than learning a single conditional policy $p_\theta(\tau_t \mid O_t)$, we define a planner-conditioned policy: $p_{\theta}(\tau_t \mid O_t,\, c_t)$, where $O_t$ is the observation context and $c_t \in \{1, \dots, M\}$ is the planner identity drawn from a finite set of $M$ planner classes. This enables a single shared model to represent a multimodal trajectory distribution, producing qualitatively different trajectories from the same observation under different planner conditions.

Each observation $o_t$ is encoded into a latent feature $z_t$ via an attention-based graph encoder following the architecture of DARE~\cite{cao2025dare}, and the resulting latents are stacked into the conditioning sequence $Z_t = (\,z_{t-T_o+1}, \dots, z_t\,)$. The planner identity $c_t$ is represented as a one-hot vector over the set of $M$ planner classes and projected to a learnable embedding $e_t$ before being fused with $Z_t$ as input to the diffusion model. The graph observation therefore provides state-dependent context capturing frontier structure and connectivity, while the planner embedding provides style-dependent context that steers the denoising process toward a particular trajectory mode. The same observation can yield qualitatively different trajectories under different planner conditions, which is the key mechanism through which the policy captures multimodality within a single shared model.

\paragraph{Diffusion training objective.}
Let $x_0$ denote the encoded clean trajectory corresponding to $\tau_t$, $x_s$ its noisy version at diffusion step $s$, and $\epsilon \sim \mathcal{N}(0, \mathbf{I})$ the injected noise. The denoising network $\epsilon_\theta$ is the noise-prediction model parameterizing the reverse diffusion process, which in turn induces $p_\theta(\tau_t \mid O_t, c_t)$. It is trained to predict the noise residual:
\begin{equation}
    \mathcal{L}_{\epsilon} = \mathbb{E}_{x_0,\,\epsilon,\,s}
    \left[\left\|\epsilon - \epsilon_{\theta}(x_s,\, s \mid Z_t,\, e_t)
    \right\|_2^2\right]
\end{equation}
 
\paragraph{Planner dropout.}
To prevent over-reliance on the planner label and encourage robustness when planner identity is unavailable, planner conditioning is randomly removed for a fraction of training samples by substituting the true label with a null condition. This regularization encourages the model to learn both planner-aware specialization and planner-agnostic exploration behavior, mirroring the classifier-free guidance technique used in conditional diffusion models~\cite{ho2021classifierfree}.

\subsection{Multi-Agent Execution and Local Reranking}
\label{sec:reranking}

During inference, the trained single-agent policy is shared across all $n$ agents. Each agent independently forms its own graph observation from the same globally shared map before sampling candidate trajectories.

\paragraph{Candidate trajectory generation.}
At each planning step, every agent samples $K$ trajectories. In the reported experiments $K = M = 8$. PCDP samples one trajectory under each of the $M$ planner conditions, whereas DARE generates $K$ unconditioned stochastic candidates; this ensures all diffusion-based methods rerank over candidate sets of equal size. Although the policy parameters are shared, trajectories differ across agents because each agent conditions on its own graph observation and its own stochastic diffusion samples. Each candidate trajectory is converted into a set of predicted traversed map cells and an estimated travel cost computed as the cumulative Euclidean length of the predicted path.

\paragraph{Coordination groups.}
Although the map is shared globally, coordination during reranking is performed locally. At each planning step, robots are partitioned into local coordination groups via a proximity graph with local radius $R_{\mathrm{local}} = 25\, \mathrm{m}$. Two robots are connected if their positions lie within this radius; connected components of this graph define the coordination groups.

\paragraph{Local reranking.}
For a coordination group $g$, $\mathcal{C}_g$ denotes the set of all joint candidate combinations formed from $K$ trajectories sampled by each robot in the group. The selected combination minimizes a local coordination score:

\begin{equation}    
    S(\mathcal{C}) = \lambda_{\mathrm{overlap}}\,\Omega(\mathcal{C}) + \lambda_{\mathrm{travel}}\,D(\mathcal{C})
\end{equation}
where $\Omega(\mathcal{C})$ is the total predicted overlap in traversed map cells across the group's trajectories and $D(\mathcal{C})$ is the sum of their travel costs; in simple terms, the robots choose the combination that works best together by reducing overlap while keeping travel cost low. In the reported experiments, $\lambda_{\mathrm{overlap}} = 1.0$ and $\lambda_{\mathrm{travel}} = 0.02$. The group reaches its joint optimum by enumerating $\mathcal{C}_g$, which costs $K^{|g|}$ score evaluations; in the worst four-agent case this gives $8^4 = 4096$ combinations.

\paragraph{Isolated robots.}
If a robot's coordination group contains only itself, reranking is not applied. That robot instead selects the best single candidate from its own $K$ samples, falling back to independent planner-guided behavior.

\paragraph{Execution.}
Once a trajectory is selected for each robot, all agents follow the receding horizon strategy of DARE: only the first action of the chosen trajectory is executed before the shared map is updated and replanning occurs. This preserves long-horizon reasoning while keeping the system reactive to newly observed information.
 
The coordination layer is deliberately lightweight: the primary source of coordination benefit is the quality and diversity of the $K$-candidate set generated by the planner-conditioned policy, while reranking is the mechanism that exploits this diversity at execution time.

\section{Results and Discussion}

We evaluate the proposed framework in a 4-agent simulation setting in which all agents share the same trained single-agent trajectory generator. Five methods are compared: two classical baselines, \textbf{Nearest}~\cite{yamauchi1998frontier} and \textbf{NBVP}~\cite{bircher2016receding}; two diffusion-only baselines, \textbf{DARE + independent} and \textbf{DARE + local reranking}; and the proposed \textbf{PCDP + local reranking}. The two DARE variants isolate the benefit of local reranking alone (DARE + local reranking vs.\ DARE + independent) and the benefit of planner conditioning under the same reranking rule (PCDP vs.\ DARE, both with local reranking). The classical baselines provide broader context on whether the proposed method remains competitive beyond diffusion-only comparisons. They are evaluated in their native multi-agent execution form on the same map instances, while the diffusion-based methods are evaluated under the shared local-reranking framework described in Section~\ref{sec:reranking}.

\subsection{Training}

All diffusion-based policies are trained on simulated environments generated by a random dungeon generator~\cite{cao2023ariadne,cao2025dare}. Each training environment is a $250 \times 250$ grid map corresponding to a $100\,\text{m} \times 100\,\text{m}$ area, with sensor range $d_s = 20\,\text{m}$ and graph node resolution $d_n = 4\,\text{m}$. All models use prediction horizon $T_p = 8$, observation horizon $T_o = 2$, and action horizon $T_a = 1$. The DARE baseline is trained on 4000 expert trajectories from a ground-truth coverage planner, following DARE~\cite{cao2025dare}. PCDP is trained on a planner-diverse dataset containing 1000 trajectories from each of $M = 8$ planner classes, giving 8000 trajectories in total.
All remaining training hyperparameters are kept identical.

\subsection{Evaluation}

Evaluation is carried out over 100 test episodes on $350 \times 350$ grid maps corresponding to $140\,\text{m} \times 140\,\text{m}$ environments, with 4 agents per episode. All methods are evaluated on the same map instances. Trajectory sampling uses DDPM inference with 100 denoising steps, and each agent draws one candidate under each of the $M = 8$ planner conditions at every planning step.

\subsection{Metrics}
We report five metrics, with $d_T^{(i)}$ denoting the total travel distance of agent $i$ at the end of an episode.
\textbf{Explored rate} is the fraction of free space discovered by the team, separating true efficiency from under-exploration.
\textbf{Success rate} is the fraction of episodes in which the team fully explores the map within the allotted episode horizon.
Three distance metrics (lower is better) are:
\textbf{Max-agent travel} $D_{\max} = \max_i\, d_T^{(i)}$ is the primary travel metric, as it captures the largest individual burden within the team;
\textbf{Total team travel} $D_{\mathrm{sum}} = \sum_i\, d_T^{(i)}$ measures aggregate motion cost; and \textbf{Agent imbalance} $D_{\mathrm{imbalance}} = \max_i\, d_T^{(i)} - \min_i\, d_T^{(i)}$ reflects how evenly work is distributed.

\subsection{Quantitative Comparison}
Table~\ref{tab:main_results} summarizes the 100-episode comparison across all
completed methods. \textbf{PCDP + local reranking} is the strongest
method overall. It preserves the same perfect success and exploration regime as the
DARE baselines, while also achieving the lowest mean max-agent travel, lowest
mean total team travel, and lowest mean agent imbalance. Relative to the
classical baselines, it also shows a substantially better efficiency--reliability
trade-off: \textbf{Nearest} reaches reasonably high exploration but at much
higher travel cost, while \textbf{NBVP} exhibits poor completion consistency and
very high imbalance.

\begin{table}[t]
    \centering
    \caption{Quantitative comparison over 100 test episodes with four agents. Travel distance metrics (lower is better) are reported as mean $\pm$ standard deviation.}
    \label{tab:main_results}
    \resizebox{\textwidth}{!}{
    \begin{tabular}{lrrrrr}
        \toprule
        Method & Success & Explored & $D_{\max}$ & $D_{\mathrm{sum}}$ &
        $D_{\mathrm{imbalance}}$ \\
        \midrule
        Nearest                     & 88\%   & 0.9618 & 982.04 $\pm$ 388.60  & 3637.21 $\pm$ 1221.60 & 158.43 $\pm$ 327.27 \\
        NBVP                        & 51\%   & 0.9554 & 1109.87 $\pm$ 210.33 & 3188.08 $\pm$ 747.41  & 685.94 $\pm$ 401.68 \\
        DARE + independent          & 100\%  & 1.0000 & 664.33 $\pm$ 81.97   & 2559.38 $\pm$ 318.53  & 49.78 $\pm$ 20.78 \\
        DARE + local reranking      & 100\%  & 1.0000 & 662.02 $\pm$ 88.55   & 2542.91 $\pm$ 346.08  & 51.37 $\pm$ 22.55 \\
        \textbf{PCDP + local reranking} & \textbf{100\%} & \textbf{1.0000} &
        \textbf{649.18} $\pm$ 91.57 & \textbf{2502.85} $\pm$ 355.91 & \textbf{48.46} $\pm$ 20.78 \\
        \bottomrule
    \end{tabular}
    }
\end{table}

\subsection{Effect of Planner Conditioning and Local Reranking}

Relative to the strongest classical baseline, \textbf{Nearest}, \textbf{PCDP + local reranking} achieves a substantially better efficiency--reliability trade-off. It improves success from 88\% to 100\%, increases mean explored rate from 0.9618 to 1.0000, and reduces mean max-agent travel by 33.9\% (982.04 $\to$ 649.18), mean total team travel by 31.2\% (3637.21 $\to$ 2502.85), and mean agent imbalance by 69.4\% (158.43 $\to$ 48.46). This shows that the proposed method does not merely maintain coverage, but does so with substantially lower travel cost and better-balanced coordination.

Against the diffusion-based baselines, the gain can be examined more directly. Compared with \textbf{DARE + local reranking}, \textbf{PCDP + local reranking} reduces mean max-agent travel by 1.94\% (662.02 $\to$ 649.18), total team travel by 1.58\% (2542.91 $\to$ 2502.85), and agent imbalance by 5.66\% (51.37 $\to$ 48.46). By contrast, the comparison between \textbf{DARE + local reranking} and \textbf{DARE + independent} shows that reranking alone yields only marginal gains: 0.35\% on max-agent travel and 0.64\% on total team travel, with imbalance slightly worsening. This indicates that planner-conditioned multimodality, rather than reranking alone, is the main contributor to improved coordination.

The gains are distributed across the team rather than concentrated in a single agent: mean travel distance is lower for all four agents under \textbf{PCDP + local reranking} than under both diffusion baselines. At the episode level, the proposed method achieves lower max-agent travel in 63 of 100 episodes against \textbf{DARE + local reranking} (59 against \textbf{DARE + independent}), and lower total team travel in 54 of 100 episodes (60 against independent). It ranks best on max-agent travel in 45 episodes and worst in only 23, indicating a consistent reduction in poor coordination outcomes rather than just an average shift.

\subsection{Qualitative Results}

Figure~\ref{fig:trajectory_diversity} shows that different planner conditions produce structurally distinct trajectory candidates from similar observations, providing qualitative support for the multimodality claim and illustrating how diverse trajectories arise from the same shared policy. Notably, trajectories generated with different planner conditions but the same graph observation diverge in both direction and extent, covering different regions. This behavioral spread is what single-planner policies lack: they tend to cluster around the same dominant mode, leaving little room for the coordination mechanism to differentiate agent assignments.

\subsection{Real-Robot Experiments}

\begin{figure}[t]
    \centering
    \hfill
    \subfigure[Within Reranking Range]
    {\includegraphics[width=0.48\textwidth]{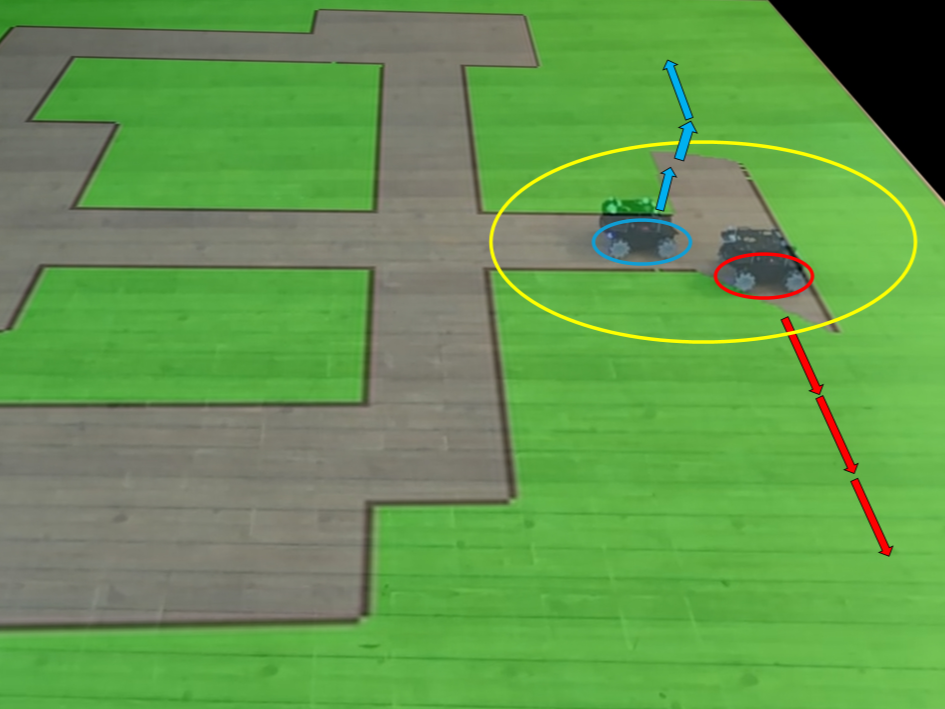}
    \label{fig:with_reranking}
    }
    \hfill
    \subfigure[Out of Reranking Range]
    {\includegraphics[width=0.48\textwidth]{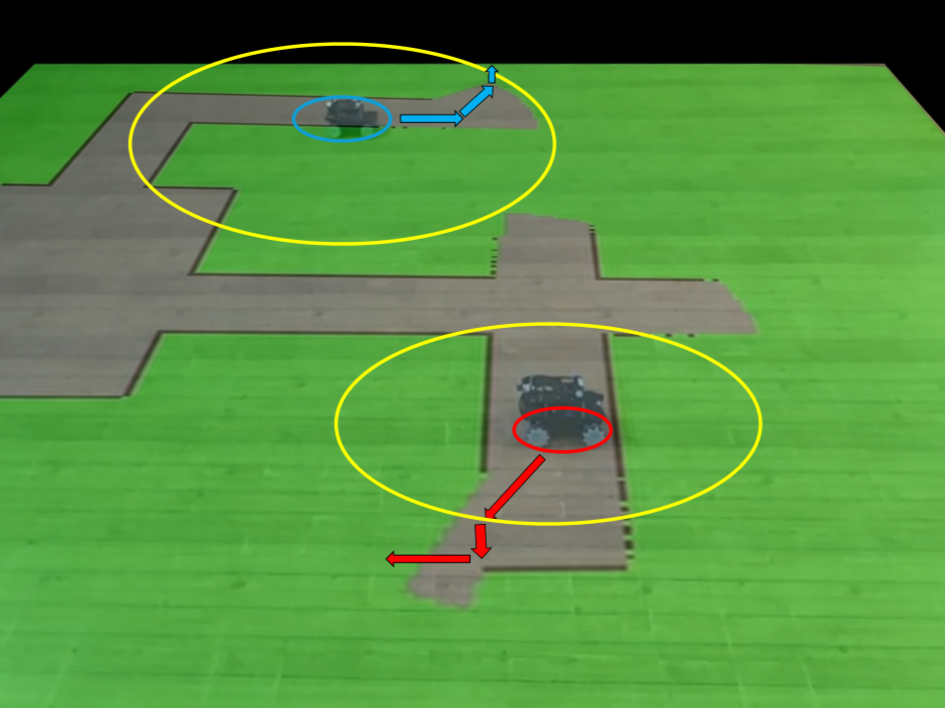}
    \label{fig:iso}
    }
    \hfill
    \caption{\textbf{Real-robot two-agent exploration under the proposed framework.}}
    \label{fig:real_robot_expt}
\end{figure}

To validate our approach beyond simulation, we deploy the framework on a two-agent platform of mecanum-wheeled robots, using the OptiTrack motion capture system for localization and a P3GASUS~\cite{p3gasus}-based centralized planning, decentralized execution framework for plan execution. The full video is included in the supplementary material.

Two complementary behaviors are observed, as illustrated in Figure~\ref{fig:real_robot_expt}. When the agents are within coordination range (Figure~\ref{fig:with_reranking}), the local reranking stage selects trajectories that diverge toward different frontiers, actively reducing redundant coverage. When an agent is isolated (Figure~\ref{fig:iso}), it falls back to independently selecting its best planner-conditioned trajectory, continuing to explore efficiently on its own. Together, these observations confirm that planner-conditioned diversity is the key enabler: without it, agents operating from similar observations would concentrate on the same dominant trajectory mode and pursue redundant paths. The framework therefore produces emergent spatial separation when agents are nearby, and effective independent exploration when they are not, without any explicit repulsion or role-assignment mechanism.

\section{Conclusion}

In this work we presented Planner-Conditioned Diffusion Policy (PCDP) for coordinated multi-agent exploration. By training a shared single-agent diffusion policy on demonstrations from multiple planner styles with planner identity as an explicit conditioning input, PCDP learns a multimodal trajectory distribution that produces diverse, controllable trajectory candidates from the same observation. Applied in a modular multi-agent framework with local reranking, the policy achieves coordination without end-to-end multi-agent training.

The central finding is that planner-conditioned multimodality, rather than the reranking mechanism alone, is the main contributor to improved coordination. Reranking over a single-planner candidate set yields only marginal gains, whereas the same rule applied to PCDP's richer candidate set reduces max-agent travel, total team travel, and agent imbalance while preserving perfect task completion. Real-robot experiments with two agents further confirm that the framework transfers to physical deployment, producing emergent spatial separation without any explicit repulsion or role-assignment mechanism.

Several limitations remain. Because PCDP uses both a larger demonstration set and greater planner diversity than DARE, the present experiments do not isolate planner conditioning from training-data volume. Local reranking also scales exponentially with coordination-group size, and the reranking objective reduces predicted overlap without enforcing inter-robot collision avoidance.

More broadly, this work demonstrates that improving the quality and diversity of the trajectory generation mechanism is a practical alternative to end-to-end multi-agent coordination learning. Future work could explore learned or adaptive reranking objectives, more efficient selection for larger coordination groups, and richer planner conditioning that further narrows the gap between single-agent training and multi-agent execution.

\subsubsection{\ackname}
This work was supported by NUS under grant TL/FS/2025/01.

\subsubsection{\discintname}
The authors have no competing interests to declare that are relevant to the content of this article.

\newpage
\bibliographystyle{splncs04}
\bibliography{ref}

@inproceedings{yamauchi1997frontier,
  title={A frontier-based approach for autonomous exploration},
  author={Yamauchi, Brian},
  booktitle={Proceedings 1997 IEEE International Symposium on Computational Intelligence in Robotics and Automation CIRA'97.'Towards New Computational Principles for Robotics and Automation'},
  pages={146--151},
  year={1997},
  organization={IEEE}
}

@inproceedings{bircher2016receding,
  title={Receding horizon" next-best-view" planner for 3d exploration},
  author={Bircher, Andreas and Kamel, Mina and Alexis, Kostas and Oleynikova, Helen and Siegwart, Roland},
  booktitle={2016 IEEE international conference on robotics and automation (ICRA)},
  pages={1462--1468},
  year={2016},
  organization={IEEE}
}

@inproceedings{cao2021tare,
  title={TARE: A Hierarchical Framework for Efficiently Exploring Complex 3D Environments.},
  author={Cao, Chao and Zhu, Hongbiao and Choset, Howie and Zhang, Ji},
  booktitle={Robotics: Science and Systems},
  volume={5},
  pages={2},
  year={2021}
}

@inproceedings{cao2023ariadne,
  title={Ariadne: A reinforcement learning approach using attention-based deep networks for exploration},
  author={Cao, Yuhong and Hou, Tianxiang and Wang, Yizhuo and Yi, Xian and Sartoretti, Guillaume},
  booktitle={2023 IEEE International Conference on Robotics and Automation (ICRA)},
  pages={10219--10225},
  year={2023},
  organization={IEEE}
}

@article{cao2024deep,
  title={Deep Reinforcement Learning-based Large-scale Robot Exploration},
  author={Cao, Yuhong and Zhao, Rui and Wang, Yizhuo and Xiang, Bairan and Sartoretti, Guillaume},
  journal={IEEE Robotics and Automation Letters},
  year={2024},
  publisher={IEEE}
}

@inproceedings{gao2018improved,
  title={An improved frontier-based approach for autonomous exploration},
  author={Gao, Wenchao and Booker, Matthew and Adiwahono, Albertus and Yuan, Miaolong and Wang, Jiadong and Yun, Yau Wei},
  booktitle={2018 15th international conference on control, automation, robotics and vision (ICARCV)},
  pages={292--297},
  year={2018},
  organization={IEEE}
}

@inproceedings{zhu2018deep,
  title={Deep reinforcement learning supervised autonomous exploration in office environments},
  author={Zhu, Delong and Li, Tingguang and Ho, Danny and Wang, Chaoqun and Meng, Max Q-H},
  booktitle={2018 IEEE international conference on robotics and automation (ICRA)},
  pages={7548--7555},
  year={2018},
  organization={IEEE}
}

@article{li2019deep,
  title={Deep reinforcement learning-based automatic exploration for navigation in unknown environment},
  author={Li, Haoran and Zhang, Qichao and Zhao, Dongbin},
  journal={IEEE Transactions on Neural Networks and Learning Systems},
  volume={31},
  number={6},
  pages={2064--2076},
  year={2019},
  publisher={IEEE}
}

@inproceedings{xu2022explore,
  title={Explore-bench: Data sets, metrics and evaluations for frontier-based and deep-reinforcement-learning-based autonomous exploration},
  author={Xu, Yuanfan and Yu, Jincheng and Tang, Jiahao and Qiu, Jiantao and Wang, Jian and Shen, Yuan and Wang, Yu and Yang, Huazhong},
  booktitle={2022 International Conference on Robotics and Automation (ICRA)},
  pages={6225--6231},
  year={2022},
  organization={IEEE}
}

@inproceedings{chiun2025marvel,
  title={MARVEL: Multi-Agent Reinforcement Learning for constrained field-of-View multi-robot Exploration in Large-scale environments},
  author={Chiun, Jimmy and Zhang, Shizhe and Wang, Yizhuo and Cao, Yuhong and Sartoretti, Guillaume},
  booktitle={2025 IEEE International Conference on Robotics and Automation (ICRA)},
  pages={11392--11398},
  year={2025},
  organization={IEEE}
}

@inproceedings{yamauchi1998frontier,
  title={Frontier-based exploration using multiple robots},
  author={Yamauchi, Brian},
  booktitle={Proceedings of the second international conference on Autonomous agents},
  pages={47--53},
  year={1998}
}

@article{burgard2005coordinated,
  title={Coordinated multi-robot exploration},
  author={Burgard, Wolfram and Moors, Mark and Stachniss, Cyrill and Schneider, Frank E},
  journal={IEEE Transactions on robotics},
  volume={21},
  number={3},
  pages={376--386},
  year={2005},
  publisher={IEEE}
}

@article{yang2023active,
  title={Active neural topological mapping for multi-agent exploration},
  author={Yang, Xinyi and Yang, Yuxiang and Yu, Chao and Chen, Jiayu and Yu, Jingchen and Ren, Haibing and Yang, Huazhong and Wang, Yu},
  journal={IEEE Robotics and Automation Letters},
  volume={9},
  number={1},
  pages={303--310},
  year={2023},
  publisher={IEEE}
}

@article{lee2024multi,
  title={Multi-Agent Exploration With Similarity Score Map and Topological Memory},
  author={Lee, Eun Sun and Kim, Young Min},
  journal={IEEE Robotics and Automation Letters},
  volume={9},
  number={11},
  pages={10327--10334},
  year={2024},
  publisher={IEEE}
}

@inproceedings{zhu2024maexp,
  title={Maexp: A generic platform for rl-based multi-agent exploration},
  author={Zhu, Shaohao and Zhou, Jiacheng and Chen, Anjun and Bai, Mingming and Chen, Jiming and Xu, Jinming},
  booktitle={2024 IEEE International Conference on Robotics and Automation (ICRA)},
  pages={5155--5161},
  year={2024},
  organization={IEEE}
}

@inproceedings{ho2020denoisingdiffusionprobabilisticmodels,
 author = {Ho, Jonathan and Jain, Ajay and Abbeel, Pieter},
 booktitle = {Advances in Neural Information Processing Systems},
 editor = {H. Larochelle and M. Ranzato and R. Hadsell and M.F. Balcan and H. Lin},
 pages = {6840--6851},
 publisher = {Curran Associates, Inc.},
 title = {Denoising Diffusion Probabilistic Models},
 url = {https://proceedings.neurips.cc/paper_files/paper/2020/file/4c5bcfec8584af0d967f1ab10179ca4b-Paper.pdf},
 volume = {33},
 year = {2020}
}

@inproceedings{
ho2021classifierfree,
title={Classifier-Free Diffusion Guidance},
author={Jonathan Ho and Tim Salimans},
booktitle={NeurIPS 2021 Workshop on Deep Generative Models and Downstream Applications},
year={2021},
url={https://openreview.net/forum?id=qw8AKxfYbI}
}

@inproceedings{janner2022diffuser,
  title = {Planning with Diffusion for Flexible Behavior Synthesis},
  author = {Michael Janner and Yilun Du and Joshua Tenenbaum and Sergey Levine},
  booktitle = {International Conference on Machine Learning},
  year = {2022},
}

@article{chi2024diffusionpolicy,
	author = {Cheng Chi and Zhenjia Xu and Siyuan Feng and Eric Cousineau and Yilun Du and Benjamin Burchfiel and Russ Tedrake and Shuran Song},
	title ={Diffusion Policy: Visuomotor Policy Learning via Action Diffusion},
	journal = {The International Journal of Robotics Research},
	year = {2024},
}

@inproceedings{cao2025dare,
  author={Cao, Yuhong and Lew, Jeric and Liang, Jingsong and Cheng, Jin and Sartoretti, Guillaume},
  booktitle={2025 IEEE International Conference on Robotics and Automation (ICRA)}, 
  title={DARE: Diffusion Policy for Autonomous Robot Exploration}, 
  year={2025},
  pages={11987-11993},
  doi={10.1109/ICRA55743.2025.11128196}}

@article{p3gasus,
  title={P3GASUS: Pre-Planned Path Execution Graphs for Multi-Agent Systems at Ultra-Large Scale},
  author={Duhan, Tanishq and He, Chengyang and Sartoretti, Guillaume},
  journal={IEEE Robotics and Automation Letters},
  volume={11},
  number={2},
  pages={1274--1281},
  year={2025},
  publisher={IEEE}
}
\end{document}